\documentclass[letterpaper]{article} 
\usepackage{aaai23}  
\usepackage{times}  
\usepackage{helvet}  
\usepackage{courier}  
\usepackage[hyphens]{url}  
\usepackage{graphicx} 
\usepackage{natbib}  
\usepackage{caption} 
\usepackage{subcaption}

\usepackage{algorithm}
\usepackage{algorithmic}

\usepackage{newfloat}
\usepackage{listings}
\DeclareCaptionStyle{ruled}{labelfont=normalfont,labelsep=colon,strut=off} 
\floatstyle{ruled}
\newfloat{listing}{tb}{lst}{}
\floatname{listing}{Listing}
\title{eCDANs: Efficient Temporal Causal Discovery from Autocorrelated and Non-stationary Data (Student Abstract)}

\author {
    Muhammad Hasan Ferdous,
    Uzma Hasan, 
    Md Osman Gani
}
\affiliations {
    Causal AI Lab, Department of Information Systems, University of Maryland, Baltimore County\\
    h.ferdous@umbc.edu, uzmahasan@umbc.edu, mogani@umbc.edu
}

\usepackage{bibentry}

\begin{document}

\maketitle

\begin{abstract}
Conventional temporal causal discovery (CD) methods suffer from high dimensionality, fail to identify lagged causal relationships, and often ignore dynamics in relations. In this study, we present a novel constraint-based CD approach for autocorrelated and non-stationary time series data (eCDANs) capable of detecting lagged and contemporaneous causal relationships along with temporal changes. eCDANs addresses high dimensionality by optimizing the conditioning sets while conducting conditional independence (CI) tests and identifies the changes in causal relations by introducing a surrogate variable to represent time dependency. Experiments on synthetic and real-world data show that eCDANs can identify time influence and outperform the baselines.
\end{abstract}

\section{Introduction}

Many substantial methods have been developed to estimate the underlying causal mechanism of time series data. But most of these approaches fail when the time series data is non-stationary and autocorrelated.
To find causal relationships in autocorrelated data, most constraint-based approaches perform conventional CI tests between variables that may include the whole past and all contemporaneous variables, from which 
some of the conditioning variables are uncorrelated 
This results in high dimensionality, lower detection power, and incorrect results \cite{pcmciPlus}. More recent works used continuous optimization to handle high-dimensionality \citep{dynotears, nts}. However, such methods can have multiple minima, can not handle data re-scaling, and the returned edges may or may not represent causal relationships \citep{unsuitability}. Moreover, the seasonal and cyclical nature of variables has a time influence that can change their distributions. This time influence is reflected in \textit{changing modules} and can be represented by a surrogate variable \cite{cdnod}. This important component of the time series data is ignored by most of the algorithms.

To address these challenges, we propose an algorithm for efficient CD from autocorrelated and non-stationary (eCDANs) data. Our proposed approach handles high dimensionality by performing PC-stable CI tests and uses the findings to perform momentary CI (MCI) tests. This gives us the adjacent sets (skeleton graph) of contemporaneous variables. The use of MCI tests ensures the exclusion of unrelated variables in conditioning sets. We then use the skeleton graph and a surrogate variable to identify the changing modules. It utilizes the direction of time flow and changing modules, and orientation rules to identify causal directions.

\section{Methodology: \texttt{eCDANs}}
Our proposed method \texttt{eCDANs} discovers causal structure in five steps. We describe these steps below.

\textbf{Step 1 -- Detection of initial adjacent sets:} Let $X_t$ be the contemporaneous variables,  \(X_{t}^{j}\) be the \(j^{th}\) observation at time t, \(X_{t-\tau}^{i}\) be the \(i^{th}\) observation at lag $\tau$. eCDANs begins by conducting iterative PC1 tests between \(X_{t}^{j}\) and \(X_{t - \tau}^{i}\) for all \(i\ (i = 1,\ 2,\ \ldots,m)\), and derives the superset of lagged adjacent set \(L_{a}(X_{t}^{j}\) ) for every \(X_{t}^{j}\). Then it performs MCI tests to get the adjacent set $Adj(X_t^j)$ for every $X_t^j$. Variables in adjacency sets are stored according to the descending effect size (test statistic).

\begin{figure*}
  \includegraphics[width=\textwidth]{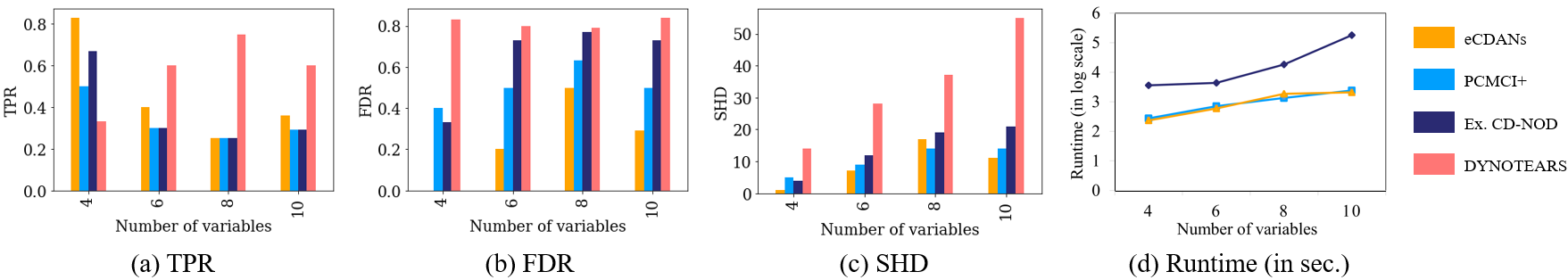}
  \caption{Performance metrics of different algorithms}
  \label{fig:comparison}
\end{figure*}

\textbf{Step 2 -- Construction of the undirected graph:} In this phase, eCDANs creates a partially complete undirected graph $G$ using the lagged adjacencies $L_{a}(X_{t})$, contemporaneous adjacencies $C_{a}(X_{t})$, and the surrogate variable $C$.

\textbf{Step 3 -- Detection of changing modules:} 
To detect changing modules, eCDANs starts with unconditional independence tests between \(X_{t}^{j}\) and \(C\) for all $j \in (1,m)$; and keeps adding other variables in the conditioning set from $(L_{a}(X_{t}^{j}) \cup C_{a}(X_{t}^{j}))$ according to the descending effect size. This optimizes the conditioning set and the search by reducing redundant conditional tests. It removes the edge between \(X_{t}^{j}\) and \(C\) if they are independent. 

\textbf{Step 4 -- Identification of contemporaneous adjacencies:} We ignored $C$ in \emph{Step--1} while detecting $C_{a}(X_{t})$ that can discover some false edges. To get rid of these edges, eCDANs performs CI tests between \(X_{t}^{i}\) and \(X_{t}^{j}\), and removes the edge between \(X_{t}^{i}\) and\(\ X_{t}^{j}\) if they are independent conditional on $(L_{a}(X_{t}^{i}) \cup C_{a}(X_{t}^{i})\cup L_{a}(X_{t}^{j}) \cup C_{a}(X_{t}^{j})\cup C)$. This step produces a causal skeleton with contemporaneous edges, lagged edges, and the edges between contemporaneous variables and $C$.

\textbf{Step 5 -- Recovery of causal direction:} Using the assumption that the cause-effect relationships follow the flow of time, it orients \({(X}_{t - \tau}^{i},\ X_{t}^{j})\) as \({(X}_{t - \tau}^{i} \rightarrow X_{t}^{j})\) for all \(\tau = 1,\ 2,\ \ldots,\ \tau_{\max}\).
eCDANs orients \(\left( C,X_{t}^{j} \right)\) as \(C \rightarrow X_{t}^{j}\) assuming that the surrogate variable causes the distribution shift. For triple of the form \(\left( C - X_{t}^{i} - X_{t}^{j} \right)\), eCDANs recalls the conditional set of the CI test between \(C\) and \(X_{t}^{j}\). If the conditioning set does not include \(X_{t}^{i}\), it orients the triple as \(C \rightarrow X_{t}^{i} \leftarrow X_{t}^{j}\). Otherwise, orients as \(C \rightarrow X_{t}^{i} \rightarrow X_{t}^{j}\). When both \(X_{t}^{i}\) and \(X_{t}^{j}\) are adjacent to \(C\), eCDANs uses extended HSIC to orient the edge between \({X}_{t}^{i}\) and \(X_{t}^{j}\). eCDANs also uses independent changes of causal modules to determine the causal direction \citep{ecdnod}.

\section{Evaluation}

We partially optimized the conditioning sets using lagged parents and benchmarked the performance of eCDANs against three baselines on -- (1) synthetic datasets (4, 6, 8, and 10 variables), and (2) a clinical dataset \cite{gani} database which contains 12 time-series variables\ref{fig:comparison}. We discuss the experimental findings below.

\textbf{Results on synthetic datasets:} Figure \ref{fig:comparison} shows the performance of eCDANs and PCMCI+ \cite{pcmciPlus}, Extended CD-NOD \cite{ecdnod}, and DYNOTEARS \cite{dynotears} in terms of True positive rate (TPR), False discovery rate (FDR), and Structural hamming distance (SHD). eCDANs consistently outperformed PCMCI+ and Extended CD-NOD in terms of all matrices. DYNOTEARS has the highest TPR, however, it fails to identify true causal relationships and produces very high FDR and SHD. Hence we did not consider DYNOTEARS for further analysis.

\textbf{Results on real-world dataset:} We can not compare the performance on the real-world dataset due to the unavailability of the ground truth causal graph. Instead, we compared the outcomes with a non-temporal causal graph proposed by \cite{gani, bikak2020}, and found that eCDANs produces the closest graph to the non-temporal findings.

\textbf{Runtime:} We present runtime for eCDANs, PCMCI+ and Extended CD-NOD in (Figure \ref{fig:comparison}). Our experimental results show that eCDANs is efficient compared to others and has the lowest runtime consistently. In the 8-variable setting, it has a slightly longer runtime than PCMCI+.

\section{Future Works}
This is a work in progress and we are working on finding the contemporaneous adjacencies along with the lagged adjacencies to further optimize the CI tests. This will further improve the performance and runtime of the proposed algorithm. We also plan to extend eCDANs when the time series dataset contains latent confounders along with changing modules to identify a partial ancestral graph. 

\section{Acknowledgements}
This study was supported in parts under grants from NSF (Award \# 2118285) and UMBC START.

\bibliography{aaai23}

\end{document}